\pdfoutput=1

\documentclass[11pt]{article}

\usepackage{acl}

\usepackage{times}
\usepackage{latexsym}

\usepackage[T1]{fontenc}

\usepackage[utf8]{inputenc}

\usepackage{microtype}

\usepackage{graphicx}
\usepackage{svg}

\usepackage{amsmath}

\title{The Fellowship of the Authors: \\ Disambiguating Names from Social Network Context}

\author{Ryan Muther \and David Smith \\
         Northeastern University \\ Boston, MA \\ muther.r@northeastern.edu, dasmith@ccs.neu.edu}

\begin{document}
\maketitle
\begin{abstract}
Most NLP approaches to entity linking and coreference resolution focus on retrieving similar mentions using sparse or dense text representations. The common ``Wikification'' task, for instance, retrieves candidate Wikipedia articles for each entity mention. For many domains, such as bibliographic citations, authority lists with extensive textual descriptions for each entity are lacking and ambiguous named entities mostly occur in the context of other named entities. Unlike prior work, therefore, we seek to leverage the information that can be gained from looking at association networks of individuals derived from textual evidence in order to disambiguate names. We combine BERT-based mention representations with a variety of graph induction strategies and experiment with supervised and unsupervised cluster inference methods. We experiment with data consisting of lists of names from two domains: bibliographic citations from CrossRef and chains of transmission (\emph{isnad}s) from classical Arabic histories. We find that in-domain language model pretraining can significantly improve mention representations, especially for larger corpora, and that the availability of bibliographic information, such as publication venue or title, can also increase performance on this task. We also present a novel supervised cluster inference model which gives competitive performance for little computational effort, making it ideal for situations where individuals must be identified without relying on an exhaustive authority list.
\end{abstract}

\section{Introduction}

\begin{figure}
\def\svgwidth{\columnwidth}
\includegraphics[width=8cm]{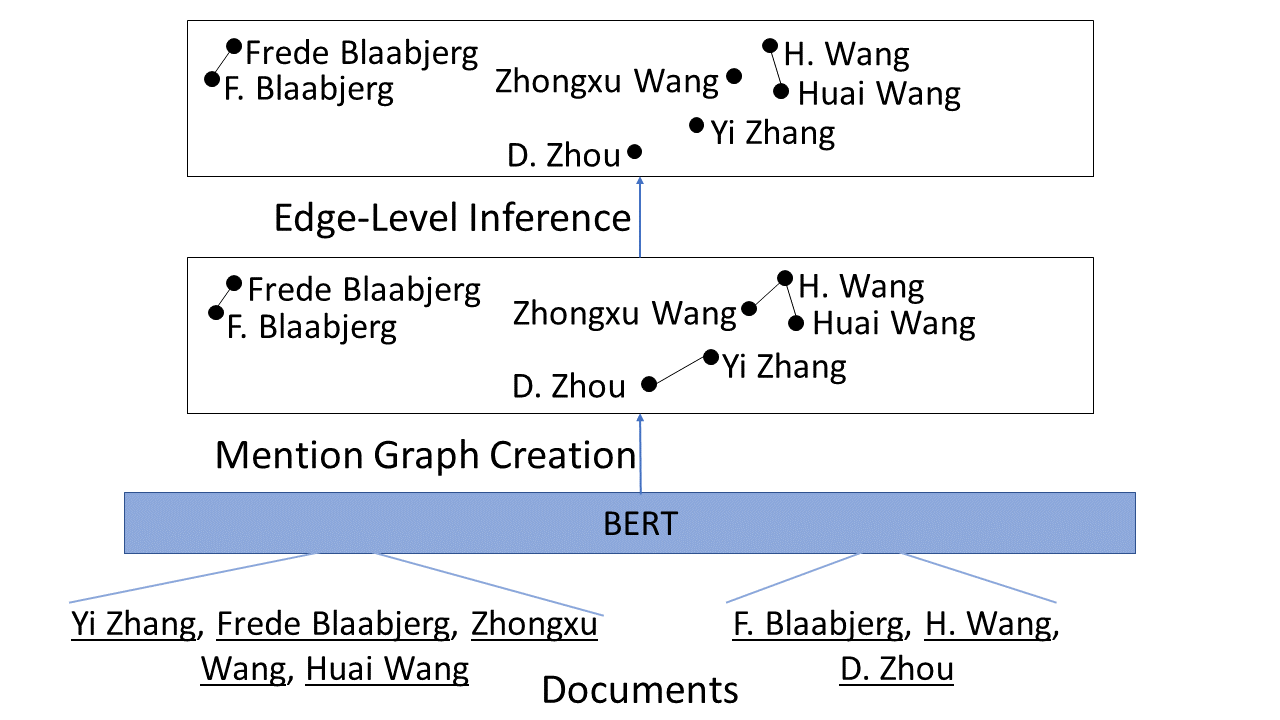}
\vspace{-8mm}
\caption{Name disambiguation using our models constructs a network of mention representations then performing inference on the edges of that network.}
\label{fig:diagram}
\end{figure}

When working with large datasets involving individuals, one is often confronted with issues of name ambiguity and resolving this ambiguity may be central to effectively using the data in downstream tasks. Take, for instance, the problem of finding all the papers authored by an individual in a database of scientific papers, such as CrossRef or arXiv. If one naively matched author names, one could either miss papers where an author published under a different name and or accidentally retrieve unrelated papers written by authors who share the name of the person of interest. Thus, more sophisticated methods must be used to determine when two names refer to the same individual. Similarly, entity linking in other genres of text like news or fiction is subject to similar constraints on how entities should be matched, with characters having multiple names or people in the news sharing names. In the more structured domain of databases, similar entity linking problems often occur in data lakes \cite{leventidis_domainnet_2021}, with names for entities occurring in varied contexts and having multiple meanings. Another potential application of these methods is in classical Arabic scholarship, where citations, rather than giving a single immediate source, provide complete a chain of transmitters (in Arabic, \textit{isnad}) and differentiating who is receiving information from whom is at the core of many questions of interest in the field. Names in \textit{isnads} are often ambiguous, with a single individual being referred to by multiple names or the same name being used for several individuals, even within the same text. While similar problems of resolving name ambiguity are often solved using entity linking systems, which take mentions and assign them to known individuals using an authority list like Wikipedia, this approach is not applicable to domains where the entities are not commonly known individuals that would show up in an authority list. To address those issues, we approach the problem of resolving name ambiguity as one of clustering, in which the goal is to infer the groups of mentions that are corfererent, removing the need for an external authority list. A general diagram of how these methods work can be seen in Figure~\ref{fig:diagram}.

One underutilized source of information that could be used to improve models of name disambiguation is the social context in which an author works. It would be be helpful, for instance, to be able to understand that two names that are dissimilar on the surface but occur in conjunction with the same set of other individuals are more likely to refer to the same person or, conversely, that two identical names in vastly different social contexts probably refer to different individuals. We leverage this information by using it to refine mention representations which are then used as the basis for a variety of supervised and unsupervised approaches to name disambiguation. In this paper, we present new datasets for name disambiguation in modern scientific papers and classical Arabic citations. We compare unsupervised network community detection baselines to a novel supervised inference method which both incorporates social network information and leverages the sequential ordering of the data to improve the quality of the inferred mention clusters. 

\section{Related Work}

The two most relevant problems to this area of research in natural language processing are entity linking and coreference resolution. Entity linking is usually done using a broad-coverage resource like Wikipedia (\citealp{durrett_joint_2014} and \citealp{mueller_effective_2018}) as an authority list with descriptions of entities and using the similarity between a mention in context and an entity’s description to assign mentions to entities, treating the problem as one of classification. However, most of the entities in our data are not in commonly-used authority lists and the context of the mentions we are working with consists almost wholly of other names, so the similarity between the mention’s context and a description of an individual in an authority list is not a useful predictor of the identity associated with a particular mention even if the correct entity were present in the authority list. 

Coreference resolution (like \citealp{lee_higher-order_2018}), in contrast, is a clustering problem in which one tries to find clusters of mentions in multiple documents that refer to the same individual. Unlike prior work on coreference resolution, our problem is slightly easier as we do not need to reason over all possible spans, as we do not have to consider cases of pronoun anaphora (like “she”) or definite descriptions (like “the actress”) as entities, only those spans that are known \textit{a priori} to be entities. In our datasets, we are either given the entity locations directly, as in CrossRef, which provides a list of authors for each paper, or we can infer their locations using a named entity recognition model, as on the \textit{isnads} (see Section 3 below). With the entity locations already known, our inference is greatly simplified compared to the more complicated coreference models in more open domains, which need to jointly infer both where entities are located and how they should be clustered to achieve strong performance. 

Also of interest is the field of entity linking in databases. \cite{leventidis_domainnet_2021}) study the problem of detecting homographs in data lakes (i.e. finding entries in a collection of database tables with the same surface form but different underlying meanings) using a co-occurrence network betweenness-centrality based scoring method. They only look at the problem of a single surface form having multiple meanings (homographs) spread across different types of items (e.g. Jaguar the car as opposed to the animal), rather than the inverse problem of multiple surface forms sharing the same meaning. They also work exclusively with the structure of the data lake itself, rather than drawing on contextual information as one could in a more text-based domain. Similar work using bibliographic \cite{bhattacharya_collective_2007} and familial relationship data \cite{kouki_collective_2019} does attempt to resolve ambiguity both in the form of homographs and synonyms, but does so in structured databases with stronger constraints, rather than in more free-form natural language.

Finally, there is the field of author name disambiguation, often studied in information science and bibliometrics, which is itself closely related to the name disambiguation problem we are solving on the CrossRef dataset. \cite{zhang_lagos-and-online-appendix_2021} propose a neural network-based model of name disambiguation which operates by classifying pairs of names as coreferent or not, but unlike our model only uses text information from the title of a publication and its abstract, ignoring the social network aspect of the name disambiguation problem. They also evaluate only the performance of their pairwise classifier, rather than looking at the quality of the clusters implied by the classifier’s results.

\section{Data and Preprocessing}

We explore network inference and name disambiguation using data from two main sources: the CrossRef paper dataset containing paper coauthorship information from the early 2000s onwards, and a collection of \textit{isnads} (chains of transmission) taken from the 12th century historical work \textit{Ta'rikh Madinat Dimashq} (History of Damascus) by Ibn ‘Asakir. Each entry in the CrossRef dataset consists of the title, journal, year of publication, and authors of a scientific paper with their associated ORCIDs (Open Researcher and Contributor ID). For the purposes of creating training data, we limit ourselves to papers for which all authors are assigned distinct ORCIDs and have a date, title, and journal. The number of documents (here, papers), individuals, and mentions in each section of the dataset can be seen below in Table~\ref{tab:crossrefStats}, using a random 60/20/20 train/dev/test split at the paper level.

\begin{table}[h]
\centering
\begin{tabular}{lllc}

& Train & Dev & Test \\
\hline
Documents & 18,584 & 6,267  & 6,296\\
\hline
Individuals & 68,861 & 25,728 & 25,800\\
\hline
Mentions & 80,710 & 27,576 & 27,697\\ 
\hline
\end{tabular}
\caption{Statistics for the CrossRef dataset.}
\label{tab:crossrefStats}
\end{table}

The vast majority of the individuals occur only once, with higher-frequency individuals quickly becoming less and less common. The most frequent individual occurs 29 times, though most of the individuals have less than 10 occurrences. A histogram of the individual frequencies can be seen in Appendix~\ref{sec:freqHistograms} 


In the \textit{isnad} dataset, each \textit{isnad} can be thought of as a long citation, giving a more complete provenance rather than just one source as is common in modern writing. Instead, \textit{isnads} give a sequence of sources, generally people, going backwards in time until a sufficiently reputable authority is reached. Rather than providing information about who was working with whom at a single point in time, like the CrossRef data, each \textit{isnad} represents the cross-temporal transmission of a story or piece of information as each person in the \textit{isnad} received information from the individual preceding it. A translated example \textit{isnad}, with transmitters underlined, can be seen in Appendix~\ref{sec:isnadExample} 

Statistics for the \textit{isnad} dataset can be seen in Table~\ref{tab:isnadStats}, again using a 60/20/20 train/dev/test split at the \textit{isnad} level. The names in the \textit{isnads} were manually disambiguated by a domain expert in the field of Arabic book history. In about ten percent of cases, she was unable to assign an identity to a name. Those mentions were left ambiguous and not included in the evaluation, save for their effect on the embeddings of disambiguated mentions. 

\begin{table}[h]
\centering
\begin{tabular}{lllc}

& Train & Dev & Test \\
\hline
Documents & 1,428 & 476  & 476\\
\hline
Individuals & 42 & 35 & 32\\
\hline
Mentions & 7,573 & 2,561 &2,554\\ 
\hline
\end{tabular}
\caption{Statistics for the \textit{isnad} dataset.}
\label{tab:isnadStats}
\end{table}

Unlike the CrossRef dataset, which is dominated by singletons, only 6 of the 44 individuals in the \textit{isnad} dataset are singletons. In contrast, the dataset is dominated by several large clusters with over one thousand mentions of some individuals. Most notable of these is the scholar Ibn Sa’d who occurs in all of the collected \textit{isnads} as a result of the domain expert’s particular research question of interest. A histogram of the individual frequencies can be seen in Appendix~\ref{sec:freqHistograms}. 


As mentioned above in Section 2 during the discussion of prior work on coreference resolution, we do not use the entity spans given to us by the annotator. Rather, to get a better picture of how effective these methods would be when applied to texts that do not come with entity annotations, the locations of entities are first inferred using a BERT-based NER model fine-tuned on a modern Arabic NER corpus \cite{obeid_camel_nodate}. Despite the linguistic mismatch between Modern Standard the resultant NER model still performs well, with a precision of .95, recall of .97, and F-score of .96.

\section{Models}

We treat the name disambiguation problem as one of clustering and experiment with two different forms of model: (1) network-based community detection and (2) a sequential model that proceeds one document at a time and uses an antecedent classifier to assign mentions to clusters with previously seen mentions as potential antecedents, building up the clusters over time. At the core of both of these models are mention representations taken from BERT-based transformer models \cite{devlin_bert_2019}. Multi-token names are represented using the unweighted average of their component token’s embeddings. For the CrossRef data, we use bert-base-cased as our base, while for the \textit{isnad} dataset we use a hybrid English-Arabic GigaBERT model from \cite{lan_empirical_2020}. As a baseline, we implement a naive network based model in which each mention is assigned to the same cluster as its nearest earlier neighbor in the network, selected by cosine distance.

In addition to using the base models, we examine the effects of tuning the name embeddings by further pretraining the models with a masked language model training objective and masking out individual names in the documents at training time. For this training, each document is converted into examples in which one name is replaced with [MASK] tokens. For instance, a paper with two authors would result in two training examples, one each with the first and second mentions masked out. The model is then trained for three epochs to predict the masked mentions given the rest of the the document, focusing on updating the embeddings of the masked mentions to improve the model’s understanding of names. 
In the experiment results below, the untuned models are referred to as “base,” while the further pretrained models are referred to as “tuned.” 

\subsection{Community Detection}

The first form of model involves constructing a network out of the mention representations, then applying community detection algorithms to find communities that correspond to distinct individuals. We experiment with two different methods of constructing networks from sets of mention embeddings; k nearest neighbors (kNN), in which each mention is connected to the k nearest mentions, and a surface form-based heuristic in which every mention is connected to every other mention that is closer than some multiplier m of the distance to the furthest identical mention. Unless otherwise noted in the experiment results, $m=1$. For singleton surface forms that have no identical mentions to define the radius, we take the average radius of all surface forms with frequency two as an estimate of the radius for frequency one. Regardless of the choice of the network creation method, we use the cosine similarity between the mention embeddings as edge weights in the network. For each dataset, we experiment with different values of k for the kNN method. In the CrossRef dataset, the small cluster sizes mean that higher values of k are likely to be ineffective. As such, we use only values of k in {5, 10, 15, 20, 25}. The \textit{isnad} dataset, in contrast, has larger clusters of coreferent names, so we experiment with values of k in increments of 5 from 5 to 100. We experiment with both label propagation \cite{raghavan_near_2007} and the Leiden algorithm \cite{traag_louvain_2019} as community detection algorithms. Label propagation is an agglomerative, iterative algorithm in which each node begins in its own cluster with a unique label. Then, each iteration, nodes are updated to the label shared by the majority of their neighbors. This process continues until all nodes have the majority label of their neighbors. The Leiden algorithm, another iterative process, also begins with a singleton partition, but proceeds to move nodes to optimize modularity, creating a new partition of the network. That partition is then used to create an aggregate network, where each element of the partition is a single node, and the node-movement and refinement process is repeated until the modularity cannot be improved. Both of these algorithms approximate optimizing the modularity of the communities.

\subsection{Antecedent Classification}

The second form of model, rather than trying to find coreferent names by looking at the entire dataset at once, assigns the mentions in the dataset to clusters iteratively working with a single document at a time using a classifier trained to select antecedents from among possible earlier mentions in the dataset. Like the neural coreference work done by (\cite{lee_end--end_2017} and \cite{lee_higher-order_2018}), we train a classifier in a similar vein to their entity similarity scoring function to determine if mentions are coreferent. Given the embedding of a mention $i$ with representation $g_i$ and potential antecedent mention $j$ with representation $g_j$, and a vector of ancillary features $\Phi(i,j)$, the model outputs whether or not mention $j$ is an antecedent (i.e. refers to the same person as) mention $i$. We define $\Phi(i,j)$ as a binary indicator variable that is 1 if the two mentions have the same surface forms, and 0 otherwise. For our antecedent classifier, we train a three-layer feedforward neural network using ReLU activation with 150 hidden units per layer and applying dropout with p=.5 at the input layer and between the linear layers to prevent overfitting. We use binary cross entropy as the loss function, giving the full loss function for a dataset of N pairs as:

\begin{align}
    -\sum_{n=1}^{N} & \left[ \nonumber y_n\,\log(\text{FFNN}(g_i,g_j,\Phi(i,j))) \,+\right.\\
    & \left. \nonumber (1-y_n)\,\log(1-\text{FFNN}(g_i,g_j,\Phi(i,j))) \right]
\end{align}

where FFNN is the probability output by the classifier for the given concatenated input mention, potential antecedent representation, and ancillary features $g_i$,$g_j$ and $\Phi(i,j)$ respectively.

This model could be used as a final series of layers on a pair of BERT models to jointly learn how to represent names and how to solve this classification problem, allowing the model to improve its mention representations for the purpose of the antecedent classification task. However, in the interest of a fair comparison with the community detection methods, we opt not to do this and hold the mention representations fixed at training time.

To construct training and test data for the antecedent classifier, we sample pairs from the mention networks constructed as described in section 4.1. For each mention in the dataset, we take the neighboring mentions that occur earlier \footnote{What “earlier” means depends on the dataset. In CrossRef, papers are ordered first by year, then by DOI, so are effectively ordered by publisher. Most papers only have years of publication, so we cannot get much more granular than that date-wise. In the \textit{isnad} dataset, each \textit{isnad} occurs at a particular location in the text. Since all the \textit{isnad} are taken from the same text, this gives a total ordering of the \textit{isnads}.} in the dataset and use their embeddings to construct training and test mention pairs. Since the networks aren't perfect, this provides both positive and negative training examples. The ratio of positive to negative examples is strongly dependent on the dataset, as well as the network creation method. The CrossRef networks, for instance, tend to have significantly more negative examples than positive examples due to the small cluster sizes, while the reverse is true for the \textit{isnad} dataset. For the CrossRef dataset, since the networks may not contain links between a mention and all their antecedents and such relationship are rare in the data, we artificially add all the possible correct antecedents to the training pair dataset to maximize the number of positive examples the model sees in training. The \textit{isnad} dataset’s networks, on the other hand, consist mostly of positive examples and sampling from them results in training sets with few negative examples, resulting in a classifier with poor precision, since it has no ability to determine when two mentions are not coreferent. To resolve this issue, we adopt an alternative sampling strategy in which the nearest 20 embeddings of earlier positive and negative examples are used to create training examples for the \textit{isnad} dataset. We will now describe the process of inferring clusterings using the antecedent classifier. 

\subsection{Cluster Inference}

Given a collection of mention representations, a network linking them, and an antecedent classifier, we infer clusters for a new document’s mentions as follows. 
For each mention, we select all the neighboring mentions that have already been clustered in the mention network as possible antecedents. We then use the antecedent classifier to filter out potential antecedents that are not classified as such. If no antecedents are predicted for a mention, that mention is assigned to a new cluster. If a mention is the only mention in that document predicted to have an antecedent in a cluster, it can be unambiguously assigned to the cluster of its most-likely predicted antecedent. If that is not the case, and multiple mentions in a document have predicted antecedents in the same cluster, we need to resolve the conflict by assigning each mention to a cluster by solving a mention-to-cluster assignment problem using the Hungarian algorithm.
For each mention, the cost for assigning that mention to a given cluster is the average likelihood of all predicted antecedents in that cluster. To disincentivize the alignment algorithm from selecting two low-likelihood assignments over one high-likelihood assignment, we allow the algorithm to opt to assign a mention to “no antecedent” by adding a additional column to the cost matrix with a score of .5 for each mention.

\section{Experiments}

We first compare using fine-tuned and untuned mention representations to understand how these models benefit from the additional information added by the representation fine-tuning process. Then, to study the usefulness of different forms of metadata in name disambiguation, we experiment with four different ways of constructing the documents used in the CrossRef dataset. As a baseline, we construct documents using only the author names separated by commas, and compare that to setups in which we prepend the paper title, the journal of publication, or both. 
For evaluation metrics, we report a selection of the metrics commonly used to evaluate models of coreference, namely MUC \cite{vilain_model-theoretic_1995}, $\text{B}^3$ \cite{bagga_entity-based_1998}, and $\text{CEAF}_e$ \cite{luo_coreference_2005}, averaged to create the CoNLL scores presented below. Since MUC and $\text{B}^3$ are inflated by the presence of a high number of singleton clusters, the results for CrossRef are given by calculating the metrics ignoring singletons in the true clusters. The results for CrossRef are reported for kNN graphs that have been thresholded to optimize CoNLL score on the development set, selecting the best algorithm and value of k also using the dev set. The selected thresholds may be found in the repository accompanying this paper. Since the threshold is so high, the kNN networks have essentially identical results regardless of the choice of k. Finally, we experiment with varying the amount of training data given to the antecedent classifier to see how well it performs relative to the unsupervised community detection models with a reduced amount of data available.

For the antecedent classifier inference method, we show results using a classifier trained on pair datasets derived from kNN networks with k=5 for the CrossRef dataset, with all possible positive examples added as described above. For the \textit{isnad} dataset, we use the alternative sampling method in which the mention network is ignored in favor of using the nearest 20 positive and negative samples for each. The main experimental results for CrossRef can be seen in Table~\ref{tab:corssrefScores} below, while Table~\ref{tab:isnadScores} shows results for the \textit{isnad} dataset.

\subsection{Effect of Representation Tuning and Network Type}

The usefulness of tuning the mention representations varies based on the dataset and the inference method of choice. The crossref dataset, as can be seen in Table~\ref{tab:corssrefScores}, generally exhibits significant increases in performance when the mention representations are fine-tuned regardless of the choice of metadata and inference method, with the only exception of using community detection with only the names or both journal and title given at the time of embedding. Generally, the effect of tuning is more pronounced for the antecedent classifier, which indicates that the embeddings of mentions with different surface forms of the same individual are improved to the point where the antecedent classifier is more capable of properly assigning them to the same cluster, resulting in an increase in recall between the untuned and tuned antecedent classifier models. We also experimented with using the surface form heuristic to construct mention networks for the CrossRef dataset, but they gave significantly worse performance so we omit their results.

\begin{table*}[t]
\centering
\begin{tabular}{llllllc}

& Base, N & Base, AC & Base, CD & Tuned, N & Tuned, AC & Tuned, CD \\
\hline
Names Only & .675 & .657 & .677 & .689 & \textbf{.841} & .692\\
\hline
+Title & .594 & .712 & .597 & .737 & \textbf{.847} & .741 \\
\hline
+Journal & .652 & .788 & .654 & .833 & \textbf{.864} & .772 \\
\hline
+Both & .824 & .713 & .816 & .719 & \textbf{.858} & .798 \\ 
\hline
\end{tabular}
\caption{CoNLL Scores for CrossRef name disambiguation models with varied metadata and embedding sources. Results given for naive (N), antecedent classifier (AC), and community detection (CD) models.}
\label{tab:corssrefScores}
\end{table*}

\begin{table*}[t]
\centering
\begin{tabular}{ccclccc}
(kNN)      & Base          & Tuned         &  & (Surface Form) & Base          & Tuned         \\ \cline{1-3} \cline{5-7} 
Naive      & .662          & .685          &  & Naive          & .753          & .797          \\ \cline{1-3} \cline{5-7} 
Leiden     & .765          & .770          &  & Leiden         & .715          & \textbf{.834} \\ \cline{1-3} \cline{5-7} 
Label Prop & .741          & .743          &  & Label Prop     & .663          & .827          \\ \cline{1-3} \cline{5-7} 
Antecedent & \textbf{.864} & \textbf{.867} &  & Antecedent     & \textbf{.829} & .814          \\ \cline{1-3} \cline{5-7} 
\end{tabular}
\caption{CoNLL Score for Base and Tuned \textit{isnad} naive, community detection, and antecedent classification models using kNN networks (left) and surface form heuristic networks (right)}
\label{tab:isnadScores}
\end{table*}

The \textit{isnad} models display less variable performance when the embeddings are tuned, with the surface form networks being the most responsive to changes in embeddings. 
It is likely that, since the topology of the surface form networks are more apt to change as the embeddings are modified, the change in embedding quality has the most effect for those networks, which in turn plays a large role in the quality of the results from community detection algorithms. The supervised antecedent classification method, on the other hand, is less sensitive to the quality both of the network and the embeddings on this dataset, which may be a side effect of the smaller variety in the texts and presence of larger clusters, making the relative locations of two embeddings of one individual’s mentions less important. The kNN networks, which judging by the worse performance of the community detection algorithms are less structurally similar to the gold standard clusters, actually give better performance when using antecedent classification compared to the surface form networks, which are more similar to the gold standard communities after tuning judging by community detection performance. A larger, more diverse \textit{isnad} dataset would show similar changes in performance to the CrossRef with respect to tuning the mention representations.

To further illustrate the importance of tuning the mention representations when using unsupervised methods, note that for CrossRef data, fine tuning increases the fraction of mentions linked to at least one antecedent from 82\% to 95\% for the kNN (k=5) dev set network, while the effect is lessened for the \textit{isnad} dataset, from 88\% to 89\% for the same setup. As such, the wider disparity in tuned and untuned performance for CrossRef
makes sense, since the changes in network structure from finetuning alter the CrossRef network in ways that improve the performance of these models more than for the \textit{isnads}. For \textit{isnads}, the antecedent classification changes little whether one tunes embeddings or not, because the classifier has learned how to determine antecedence regardless of the presence of the additional linguistic information added by the fine tuning process. The small size of the dataset may have caused the embeddings to not shift enough for the performance to significantly change.

To understand how much the performance depends on community detection algorithms rather than on network structure alone, we compare against a naive baseline in which each mention is assigned to the same cluster as its nearest earlier neighbor in the network. This naive method
cannot assign mentions to singleton clusters unless a mention is disconnected from the rest of the network. Comparing the results for this method to those for community detection using the Leiden algorithm on the CrossRef dataset with added Journal metadata (Table~\ref{tab:comparison}), it can be seen that there is a clear precision-recall tradeoff. The naive model still has high precision, so the structure of the network contributes a fair amount of performance, properly isolating distinct clusters from the rest of the network, implying that the network on its own contributes some information rather than being reliant on the quality of the community detection algorithm on its own. However, expanding the evaluation to include singleton mentions makes the precision-recall tradeoff much more obvious. Consequently, the simple cluster assignment model's inability to assign mentions to singleton clusters causes its precision to suffer in the easy cases, where two singleton names happen to be nearest to each other in the network but are completely unrelated.  Community detection algorithms like Leiden can resolve those issues by not clustering two mentions together if they are connected by a low-weight edge, though such a method can occasionally overdo this, as the decreased MUC recall for community detection shows. The simple inference model, as one would expect, struggles with precision but does better on recall. $\text{B}^3$ scores are of course still generally high as there are many singletons and larger clusters that the naive model can do quite well on.

\begin{table}[h]
\centering
\begin{tabular}{ll}
Method & CoNLL Score \\ \hline
Naive  & .833        \\ \hline
CD     & .722       
\end{tabular}
\caption{Evaluation of Leiden community detection on the CrossRef dataset compared to a naive nearest-neighbor baseline.}
\label{tab:comparison}
\end{table}

A similar comparison with the \textit{isnad} dataset shows a comparable, albeit lower, tradeoff with the recall of the naive model being slightly higher and precision slightly lower. MUC is less obviously affected as the dataset contains fewer singletons.

\subsection{Effect of CrossRef Metadata}

Table~\ref{tab:corssrefScores} also shows the effect of including additional context in the language model when computing name embeddings. These metadata fields, such as paper title or journal name, are appended to the list of author names as they would be in a plain text citation. With the anomalous exception of untuned community detection, adding the title alone or both the title and journal helps slightly, while the journal alone gives the highest increase in performance. It is possible that adding the journal alone is more effective than adding both due to the dissimilarity in titles making the embeddings less useful, as scholars may try to make their titles unique, forcing mentions of their name further apart in embedding space. Additionally, one should note that the performance variance across different metadata settings decreases when one uses tuned embeddings or supervised methods.

\subsection{Effect of Training Data Quantity}

To understand how much benefit the additional training data and annotation time brings, we investigate the rate at which the inference performance improves as more documents are used to train the antecedent classifier. The plots used for this analysis can be seen in Appendix~\ref{sec:learningCurves}. On the \textit{isnad} dataset, the scores stay in the high 80s, with peak performance at using only 40\% of the training data. This is likely an artifact of the small size of the dataset. It is possible that that 40\% of the training data was more similar to the test set, and as the additional training data was added, the more general model started to struggle on the test set in comparison. The CrossRef results are more like what one would expect, with score increasing with training data quantity. The supervised model does not surpass community detection until 40\% of the data is used for training, likely because of the higher diversity of the datatset.

\section{Discussion}

There are several takeaways from these experiments for those interested in solving similar problems of name ambiguity in low external-resource domains. Firstly, depending on the size of the dataset, unsupervised community detection methods may give competitive performance when compared to more complex supervised inference methods. If one opts to use unsupervised methods, the quality of one’s mention representations is extremely important, and should be improved by additional pretraining of the source language model. For supervised methods, the effect of further pretraining depends more on the size of the dataset in question, with larger, more diverse datasets benefiting more from pretraining. Similarly, depending on the distribution of cluster sizes and the division of surface forms between clusters, different network construction heuristics will give different clustering results.
As $k$ increases for the CrossRef dataset, performance tends to decrease as the network includes irrelevant edges, even with aggressive pruning. The \textit{isnads}, in contrast, display increased performance as $k$ increases, likely because for any given mention, the distinct surface forms of the individual that mention refers to are going to be further from mentions of the same surface form, and thus the connections to those other regions of embedding space do not exist in the low-$k$ networks. Using the surface form heuristic, we find that even for the \textit{isnad} data, where the method performed well, performance begins to drop off quickly as the multiplier increases, with the optimal multiplier never being greater than 1.2. When the surface forms are spread across a wide variety of clusters and occurring in many varied contexts, like in CrossRef, the kNN networks become more useful, while the surface form networks perform better when each surface form is more associated with a single individual. Additional metadata associated with the mentions, like the bibliographic information provided by CrossRef, is also useful as it allows otherwise distinct mentions to be moved closer together in embedding space. Finally, for larger datasets, if annotating data from scratch, one would need to do a fair bit of annotation for the kind of supervised models we explore here to surpass the performance of unsupervised models, but once one gets past a certain point there will be diminishing performance returns from the addition of more training data. 

This work also serves as the first step in future work exploring this problem of name ambiguity in more detail, whether in other domains or in expanded versions of the current datasets, like expanding the \textit{isnad} dataset to cover multiple texts and dealing with the issues that would arise from varied authorial style, or integrating the partially annotated papers present in the CrossRef dataset. Related inference problems in these networked texts, such as inferring missing nodes or estimating the number of individuals present in a dataset, could also be of interest. One could also use the output of these models as the beginning of further work in social network analysis. The networks created by these sorts of models would also be invaluable for understanding the roles played by individuals in these social networks of knowledge production.

\section*{Acknowledgements}

Omitted for peer review

\bibliography{custom}

\begin{thebibliography}{16}
\expandafter\ifx\csname natexlab\endcsname\relax\def\natexlab#1{#1}\fi

\bibitem[{Bagga and Baldwin(1998)}]{bagga_entity-based_1998}
Amit Bagga and Breck Baldwin. 1998.
\newblock \href {https://doi.org/10.3115/980845.980859} {Entity-{Based}
  {Cross}-{Document} {Coreferencing} {Using} the {Vector} {Space} {Model}}.
\newblock In \emph{36th {Annual} {Meeting} of the {Association} for
  {Computational} {Linguistics} and 17th {International} {Conference} on
  {Computational} {Linguistics}, {Volume} 1}, pages 79--85, Montreal, Quebec,
  Canada. Association for Computational Linguistics.

\bibitem[{Bhattacharya and Getoor(2007)}]{bhattacharya_collective_2007}
Indrajit Bhattacharya and Lise Getoor. 2007.
\newblock \href {https://doi.org/10.1145/1217299.1217304} {Collective entity
  resolution in relational data}.
\newblock \emph{ACM Transactions on Knowledge Discovery from Data}, 1(1):5.

\bibitem[{Devlin et~al.(2019)Devlin, Chang, Lee, and
  Toutanova}]{devlin_bert_2019}
Jacob Devlin, Ming-Wei Chang, Kenton Lee, and Kristina Toutanova. 2019.
\newblock \href {http://arxiv.org/abs/1810.04805} {{BERT}: {Pre}-training of
  {Deep} {Bidirectional} {Transformers} for {Language} {Understanding}}.
\newblock \emph{arXiv:1810.04805 [cs]}.
\newblock ArXiv: 1810.04805.

\bibitem[{Durrett and Klein(2014)}]{durrett_joint_2014}
Greg Durrett and Dan Klein. 2014.
\newblock \href {https://doi.org/10.1162/tacl_a_00197} {A {Joint} {Model} for
  {Entity} {Analysis}: {Coreference}, {Typing}, and {Linking}}.
\newblock \emph{Transactions of the Association for Computational Linguistics},
  2:477--490.

\bibitem[{Kouki et~al.(2019)Kouki, Pujara, Marcum, Koehly, and
  Getoor}]{kouki_collective_2019}
Pigi Kouki, Jay Pujara, Christopher Marcum, Laura Koehly, and Lise Getoor.
  2019.
\newblock \href {https://doi.org/10.1007/s10115-018-1246-2} {Collective entity
  resolution in multi-relational familial networks}.
\newblock \emph{Knowledge and Information Systems}, 61(3):1547--1581.

\bibitem[{Lan et~al.(2020)Lan, Chen, Xu, and Ritter}]{lan_empirical_2020}
Wuwei Lan, Yang Chen, Wei Xu, and Alan Ritter. 2020.
\newblock \href {http://arxiv.org/abs/2004.14519} {An {Empirical} {Study} of
  {Pre}-trained {Transformers} for {Arabic} {Information} {Extraction}}.
\newblock \emph{arXiv:2004.14519 [cs]}.
\newblock ArXiv: 2004.14519.

\bibitem[{Lee et~al.(2017)Lee, He, Lewis, and Zettlemoyer}]{lee_end--end_2017}
Kenton Lee, Luheng He, Mike Lewis, and Luke Zettlemoyer. 2017.
\newblock \href {http://arxiv.org/abs/1707.07045} {End-to-end {Neural}
  {Coreference} {Resolution}}.
\newblock \emph{arXiv:1707.07045 [cs]}.
\newblock ArXiv: 1707.07045.

\bibitem[{Lee et~al.(2018)Lee, He, and Zettlemoyer}]{lee_higher-order_2018}
Kenton Lee, Luheng He, and Luke Zettlemoyer. 2018.
\newblock \href {http://arxiv.org/abs/1804.05392} {Higher-order {Coreference}
  {Resolution} with {Coarse}-to-fine {Inference}}.
\newblock \emph{arXiv:1804.05392 [cs]}.
\newblock ArXiv: 1804.05392.

\bibitem[{Leventidis et~al.(2021)Leventidis, Rocco, Gatterbauer, Miller, and
  Riedewald}]{leventidis_domainnet_2021}
Aristotelis Leventidis, Laura~Di Rocco, Wolfgang Gatterbauer, Renée~J. Miller,
  and Mirek Riedewald. 2021.
\newblock \href {https://arxiv.org/abs/2103.09940} {{DomainNet}: {Homograph}
  {Detection} for {Data} {Lake} {Disambiguation}}.
\newblock \emph{CoRR}, abs/2103.09940.
\newblock ArXiv: 2103.09940.

\bibitem[{Luo(2005)}]{luo_coreference_2005}
Xiaoqiang Luo. 2005.
\newblock \href {https://aclanthology.org/H05-1004} {On {Coreference}
  {Resolution} {Performance} {Metrics}}.
\newblock In \emph{Proceedings of {Human} {Language} {Technology} {Conference}
  and {Conference} on {Empirical} {Methods} in {Natural} {Language}
  {Processing}}, pages 25--32, Vancouver, British Columbia, Canada. Association
  for Computational Linguistics.

\bibitem[{Mueller and Durrett(2018)}]{mueller_effective_2018}
David Mueller and Greg Durrett. 2018.
\newblock \href {https://doi.org/10.18653/v1/D18-1126} {Effective {Use} of
  {Context} in {Noisy} {Entity} {Linking}}.
\newblock In \emph{Proceedings of the 2018 {Conference} on {Empirical}
  {Methods} in {Natural} {Language} {Processing}}, pages 1024--1029, Brussels,
  Belgium. Association for Computational Linguistics.

\bibitem[{Obeid et~al.(2020)Obeid, Zalmout, Khalifa, Taji, Oudah, Alhafni,
  Inoue, Eryani, Erdmann, and Habash}]{obeid_camel_nodate}
Ossama Obeid, Nasser Zalmout, Salam Khalifa, Dima Taji, Mai Oudah, Bashar
  Alhafni, Go~Inoue, Fadhl Eryani, Alexander Erdmann, and Nizar Habash. 2020.
\newblock {CAMeL} {Tools}: {An} {Open} {Source} {Python} {Toolkit} for {Arabic}
  {Natural} {Language} {Processing}.
\newblock \emph{Proceedings of the 12th Language Resources and Evaluation
  Conference}, page~11.

\bibitem[{Raghavan et~al.(2007)Raghavan, Albert, and
  Kumara}]{raghavan_near_2007}
Usha~Nandini Raghavan, Réka Albert, and Soundar Kumara. 2007.
\newblock \href {https://doi.org/10.1103/PhysRevE.76.036106} {Near linear time
  algorithm to detect community structures in large-scale networks}.
\newblock \emph{Physical Review E}, 76(3):036106.

\bibitem[{Traag et~al.(2019)Traag, Waltman, and van Eck}]{traag_louvain_2019}
V.~A. Traag, L.~Waltman, and N.~J. van Eck. 2019.
\newblock \href {https://doi.org/10.1038/s41598-019-41695-z} {From {Louvain} to
  {Leiden}: guaranteeing well-connected communities}.
\newblock \emph{Scientific Reports}, 9(1):5233.

\bibitem[{Vilain et~al.(1995)Vilain, Burger, Aberdeen, Connolly, and
  Hirschman}]{vilain_model-theoretic_1995}
Marc Vilain, John Burger, John Aberdeen, Dennis Connolly, and Lynette
  Hirschman. 1995.
\newblock \href {https://doi.org/10.3115/1072399.1072405} {A model-theoretic
  coreference scoring scheme}.
\newblock In \emph{Proceedings of the 6th conference on {Message} understanding
  - {MUC6} '95}, page~45, Columbia, Maryland. Association for Computational
  Linguistics.

\bibitem[{Zhang(2021)}]{zhang_lagos-and-online-appendix_2021}
Li~Zhang. 2021.
\newblock \href {https://doi.org/10.5281/ZENODO.4659685}
  {{LAGOS}-{AND}-{Online}-{Appendix}}.
\newblock \emph{CoRR}.
\newblock Publisher: Zenodo Version Number: 1.0.

\end{thebibliography}
\bibliographystyle{acl_natbib}

\newpage
\appendix

\section{Individual Frequency Histograms}
\label{sec:freqHistograms}

\begin{figure}[h]
\includegraphics[width=8cm]{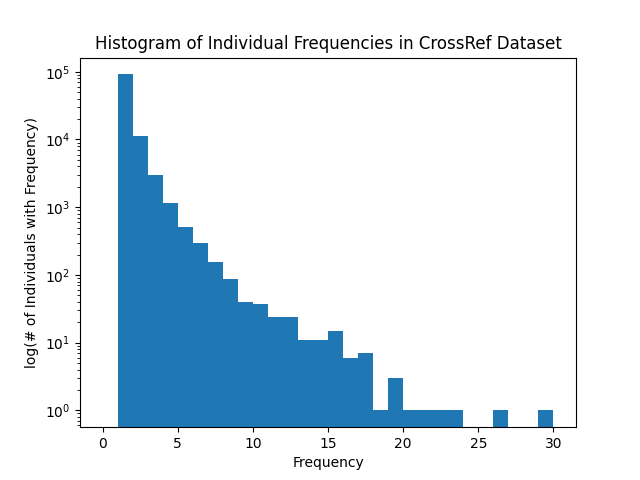}
\caption{Histogram of log(individual frequencies) in the CrossRef dataset.}
\label{fig:crossrefHist}
\end{figure}

\begin{figure}[h]
\includegraphics[width=8cm]{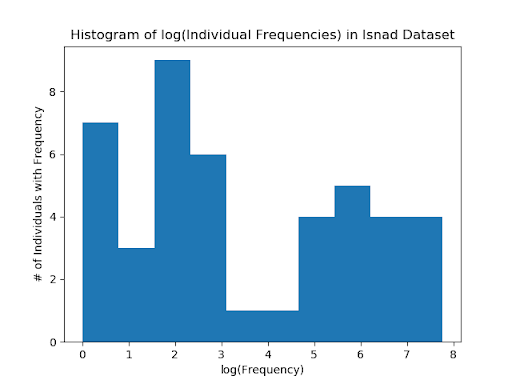}
\caption{Histogram of log(individual frequencies) in the \textit{isnad} dataset (The [0,1) bin has 7 items as one individual has a low enough frequency to be binned with the frequency 1 individuals)}
\label{fig:isnadHist}
\end{figure}

Figures~\ref{fig:crossrefHist} and \ref{fig:isnadHist} show the frequencies of individuals in the CrossRef and \textit{isnad} datasets respectively to accompany the description given of the cluster size distributions given in Section 3

\section{Example Isnad}
\label{sec:isnadExample}


Although we use the original Arabic texts of \textit{isnads} in the experiments, this English translation gives some idea of the typical brevity of each name reference:

\begin{quote}
\underline{Abu Dawud} transmitted to us, saying, 
\underline{‘Hisham} transmitted to us, from \underline{Qatadah}, from \underline{al-Hasan}, from \underline{Samurah} that the Prophet, may the peace and blessing of God be on him (al-Tayalisi, 204AH/819CE)
\end{quote}

\section{CoNLL Score as a Function of Training Corpus Size}
\label{sec:learningCurves}

\begin{figure}[h]
\includegraphics[width=8cm]{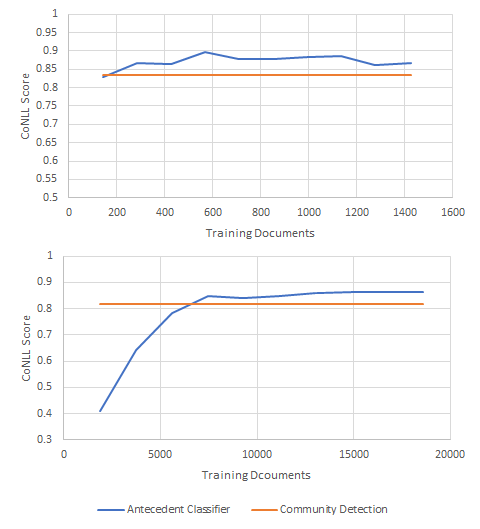}
\caption{CoNLL Score as a function of training corpus size for antecedent classification models for isnads (top) and CrossRef (bottom) compared to the score of the best-performing community detection models}
\label{fig:crossCurves}
\end{figure}

Figure~\ref{fig:crossCurves} shows a plot of CoNLL score versus training corpus size for both datasets used in the analysis in Section 5.3

\end{document}